# Biohybrid Microrobots Based on Jellyfish Stinging Capsules and Janus Particles for In Vitro Deep-Tissue Drug Penetration

Sinwook Park[1, 3], Noga Barak[2], Tamar Lotan[2], and Gilad Yossifon[1, 3*]

[1]School of Mechanical Engineering, Tel-Aviv University, Tel Aviv, 6997801, Israel

[2]Marine Biology Department, The Leon H. Charney School of Marine Sciences, University of Haifa, Haifa, 3103301, Israel

[3]Department of Biomedical Engineering, Tel-Aviv University, Tel Aviv, 6997801, Israel

* Corresponding author: gyossifon@tauex.tau.ac.il

## Abstract

Microrobots engineered from self-propelling active particles, extend the reach of robotic operations to submillimeter dimensions and are becoming increasingly relevant for various tasks, such as manipulation of micro/nanoscale cargo, particularly targeted drug delivery. However, achieving deep-tissue penetration and drug delivery remain a challenge. This work developed a novel biohybrid microrobot consisting of jellyfish stinging capsules, which act as natural nanoinjectors for efficient penetration and delivery, assembled onto an active Janus particle (JP). While microrobot transport and navigation was externally controlled by magnetic field-induced rolling, capsule loading onto the JP surface was controlled by electric field. Following precise navigation of the biohybrid microrobots to the vicinity of target tissues, the capsules were activated by a specific enzyme introduced to the solution, which then triggered tubule ejection and release of the preloaded molecules. Use of such microrobots for penetration of and delivery of the preloaded drug/toxin to targeted cancer spheroids and live *Caenorhabditis elegans* was demonstrated in-vitro. The findings offer insights for future development of bio-inspired microrobots capable of deep penetration and drug delivery. Future directions may involve encapsulation of various drugs within different capsule types for enhanced versatility. This study may also inspire in-vivo applications involving deep tissue drug delivery.

## Introduction

Active particles, often referred to as micromotors or microswimmers, self-propel under uniform ambient conditions such as magnetic, electric, or chemical fields.[1] When combined with closed-loop controlled navigation, they can be regarded as microrobots (MRs), which are gaining recognition as precise and controllable tools with numerous potential medical applications, including targeted drug delivery, micro-surgery and biomedical diagnosis[2,3]. Despite intensive study on the use of micro-nano-robots (MNRs) for targeted drug delivery, achieving deep tissue penetration and effective drug delivery remain significant challenges.

Previous studies on tissue penetration of MRs predominantly relied on synthetic MRs externally powered by a magnetic field or acoustic propulsion[4–13]. Magnetically powered microrobots bring the advantages of fuel-free actuation, remote path control and programmability[4–10]. For example, a magnetically powered microdriller[9] and a magnetic propeller[8] both with corkscrew-type rotating motion, have been used to penetrate gels and the vitreous body of the eye[8]. Wang et al. presented acoustic-powered MRs that utilize ultrasound pulses to initiate droplet vaporization of encapsulated biocompatible fuel, resulting in high-velocity, bullet-like propulsion of drug-carrying nanoparticles preloaded within the MR[11,12]. These MRs have been shown to penetrate hard tissues, e.g. liver. Yet, while acoustic-powered MRs exhibit deeper tissue penetration and delivery of nanoparticle payload than that of magnetically powered MRs, the penetration depth into a tissue gel matrix was limited to ~20 μm[12].

This study presents a novel drug delivery approach which harnesses jellyfish nematocysts as natural nanoinjectors for efficient deep penetration into tissues. Nematocysts, the stinging capsules of jellyfish, are inherently explosive, natural-injection systems with significant potential as a drug-delivery system[14–17]. They are comprised of a capsule containing a compactly folded, needle-like microtubule and a highly concentrated matrix with charged constituents that maintain an extremely high inner osmotic pressure of ~150 bar which facilitates tubule firing and penetration into its target[18–20]. The permeable, negative charged wall of the jellyfish capsule enables accumulation of molecules within the capsule body, particularly those of sizes up to 600 Da[21–24]. During the discharge process, the tubule contained within the capsules undergoes an inside-out eversion at an acceleration rate of ~$5 \times 10^6$ g and extends to over 100-times the capsule's diameter, enabling it to penetrate into relatively rigid target materials[25,26]. Subsequently, molecules preloaded within the capsule are ejected through the tubule's generally porous surface, primarily through its tip opening[17].

The formation of biohybrid MRs consisting of jellyfish capsules can be achieved through the strategic dielectrophoretic pairing (DEP) of synthetic active particles (e.g., Janus particles, JP) with intact jellyfish capsules with encapsulated payloads. Such a combination offers devices with synergistic behavior of naturally occurring mechanisms with engineered functionalities. For example, biohybrid micromotors and microrobots were recently engineered for drug delivery into hard cancer tissues by coupling motile microorganisms (e.g., bacteria[27–29], sperm cell[30]) that swim via flagellar motion, with drug-loaded synthetic nanoparticles[31]. Our recent investigations utilizing metallo-dielectric JPs demonstrated electrically unified control over both self-propulsive motion and DEP-based cargo manipulation[32–35]. JPs function as micromotors and cargo carriers, and under the application of an electric field, were recently shown to transport antibody-functionalized beads for sandwich immunoassay-based biosensing[33], as well as localized electroporation and transfection of drugs/genes into cells[36]. This work aimed to assemble biohybrid MRs consisting of synthetic JPs and natural jellyfish capsules and to characterize their transport under combined magnetic and electric field actuations. It also assessed the potential of navigating them to targeted regions within a microfluidic chamber and to perform in-vitro injection of the molecular contents of the jellyfish capsules deep into targeted cancer spheroids and a *C. elegans* model.

## Results and Discussion

### Assembly of biohybrid microrobots and characterization of their transport

An innovative strategy was used to develop biohybrid MRs for precise transport and activation of jellyfish capsules preloaded with drugs (Fig. 1a). The micromotor, consisting of JPs, exhibits controlled self-propulsion and navigation through a magnetic rolling field, along with an external alternating (AC) electric field[37]. The micromotors and the jellyfish capsules are assembled by dielectrophoretic (DEP) trapping forces under an external AC electric field. Once the biohybrid MRs are formed, magnetic rolling is primarily employed to transport them to the targeted region. Activation of the jellyfish capsules involves the introduction of an enzyme (1% subtilisin protease) that destabilizes the operculum of the capsule, resulting in an osmotic pressure difference, followed by rapid tubule discharge[20]. While subtilisin protease was chosen for convenience, other enzymes and compounds such as ethylene glycol tetraacetic acid (EGTA), methylene blue, and various salts are also alternatives.[22,38] Fig.1b depicts a biohybrid MR formed (I) by combining a 27μm JP with ten jellyfish capsules within a PBS (σ ~ 18 mS cm$^{-1}$) solution inside a microchamber. The MR was then transported to the targeted region (II) via magnetic field-based rolling, coupled with AC field-based orientation (5 MHz, 15$V_{pp}$), and tubules were ejected (III) upon interaction with the enzyme.

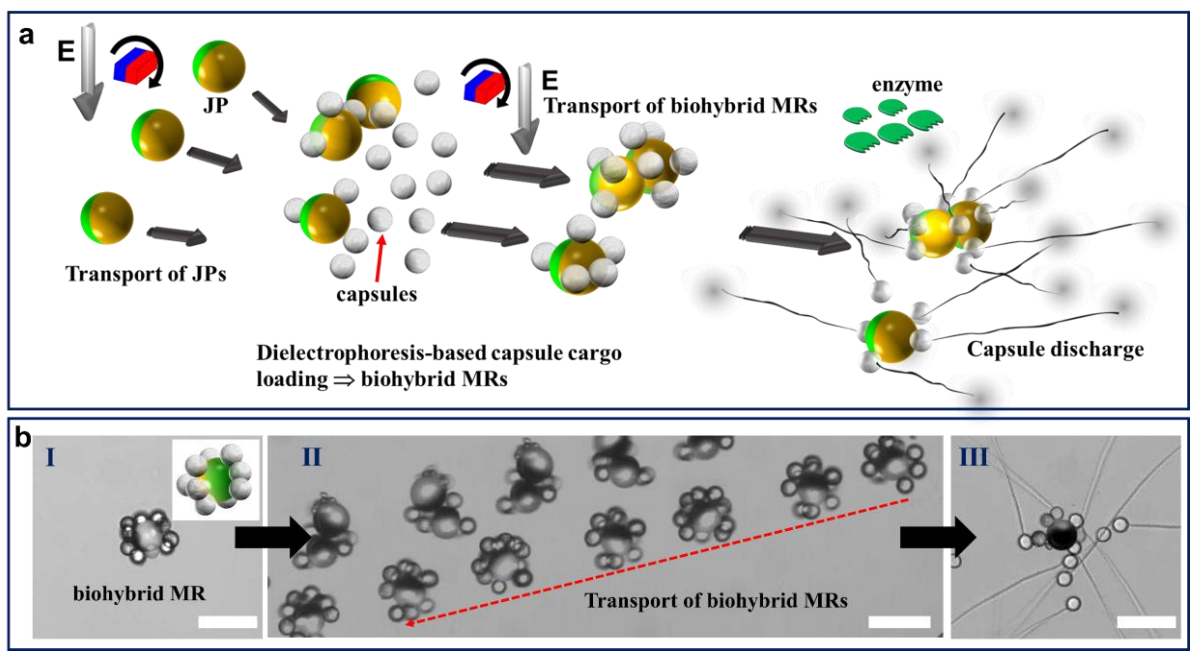

**Figure 1. Biohybrid microrobots comprised of Janus particles and jellyfish capsules.** (a) Conceptual illustration of the biohybrid microrobots (MRs) comprised of jellyfish capsules as cargo on metallo-dielectric Janus particle (JP) carriers. The cargo-loaded JPs are transported via magnetic rolling, while an electric field aligns

the JPs and traps the capsules in regions with the activating enzyme, which triggers capsule discharge with rapid microtubule ejection in various directions. The metallic coating (Cr/Ni/Au) on the JP is depicted in yellow, while the JPs' bare polystyrene hemisphere is depicted in green. (b) Time-lapse microscopic images depicting (I) a representative biohybrid MR loaded with more than ten capsules within a PBS solution ($\sigma \sim 18$ mS cm$^{-1}$), (II) the transport of biohybrid MRs using magnetic rolling with an applied electric field and (III) capsule discharge triggered by the enzyme. The image in II was generated by superimposing images taken at two-second intervals. The red arrow indicates the direction of propulsion. The magnetic field rotation was 100 rpm, and the applied electric field was 10 kHz, with 15V$_{pp}$. Scale bar: 50μm.

To optimize the trapping of jellyfish capsules and the transportation of assembled MRs, the dynamics of the JPs and their cargo-loading behaviors were investigated under varying electric and magnetic fields and solution conductivities (Fig. 2). This included the application of either only electric field for singular control of loading and transport or a combination of magnetic rolling and an electric field. Prior to capsule trapping, we examined the transport behavior of JPs in different concentrations of NaCl solutions from 0.01 to 10 mM NaCl under three conditions: magnetic rolling only, electric field only, coupled magnetic rolling and electric field (Fig S1 and movie S2). Under magnetic rolling, the velocity of JPs slightly increased with higher NaCl concentrations due to the decreased electric double layer (EDL) thickness, which increased the effective friction between the JPs and the bottom substrate. When driven by an electric field only, the JP's self-propulsion exhibited either induced charge electrophoretic (ICEP) or self-dielectrophoretic (sDEP) propulsion depending on the applied frequency.[32–35] Increasing the frequency of the AC electric field resulted in a transition from ICEP (JP moves with its dielectric side forward) to sDEP (JP moves with its metallic-coated side forward) propulsion. As the NaCl concentration increased, the velocities of JPs in both ICEP and sDEP modes decreased, reaching nearly zero at 10 mM NaCl. Such decrease in the magnitude of the electrokinetic velocity with increasing solution conductivity is well known and still not fully understood as it necessitates theories that go beyond the linearized electric double layer (EDL) descriptions[39]. In particular, as the RC time of the induced EDL is decreasing with increasing solution conductivity (as the Debye length scales inversely with the square root of the solution conductivity), the frequencies (inversely proportional to the RC time) for transitioning between ICEP and sDEP are also shifted to higher values[39]. Combining magnetic rolling with an electric field resulted in the superposition of both electric-field driven and magnetic rolling vectors, which adjusted the angles according to the applied frequencies. This mode, particularly at higher frequencies (>2 MHz), enhanced the magnetic rolling-induced velocity and controllability due to the electric-field induced orientation, especially at concentrations above 10 mM NaCl, as explained in our previous work[37]. Additionally, the DEP responses of jellyfish capsules in varying NaCl concentrations were comprehensively characterized (Fig. S2). At NaCl concentrations below 0.1 mM and above 10 mM, the capsules exhibited positive dielectrophoresis (pDEP) and negative dielectrophoresis (nDEP) behavior, respectively, across all frequency ranges. At the intermediate solution concentration of 1 mM, the DEP response shifted from nDEP to pDEP at a cross-over frequency between 100 kHz and 500 kHz. Fig. 2a and movie S1 illustrate the representative transport and on-the-fly cargo loading of a biohybrid MR propelled by a unified external AC electric field (2 kHz, 15 $V_{pp}$) in the presence of 0.1 mM NaCl. Under these conditions, the JP propelled due to ICEP motion, with its dielectric hemisphere forward, while trapping capsules on its equator due to a pDEP response. The maximum number of trapped capsules reached 4 to 5 at the dielectric hemisphere's equator, with even more capsules trapped in other locations under solely electric field application (e.g., 1 kHz, 0.1 mM NaCl, Fig. S4a) without magnetic rolling. However, in ICEP mode, the propulsion direction of multiple JPs appeared random due to the initial orientation of their dielectric surfaces. In sDEP mode, JPs tended to be stuck to the bottom surface by electrostatic attraction, making transport difficult (Fig. S1). Thus, magnetic rolling or steering was necessary for precise transport of assembled microbots in both cases.

The combination of an electric field and magnetic rolling enabled controllable transport of JPs for capsule trapping. Fig. 2c shows representative pDEP and nDEP trapping of the capsules on a single JP during the transport at varying NaCl conductivities under the tested conditions with various frequencies. Increasing the frequency resulted in a shift between nDEP and pDEP responses of capsules at the intermediate solution conductivity of 1 mM NaCl, while keeping pDEP based capsule trapping within solution conductivities below 0.1 mM NaCl (Fig. 2c, Fig. S2 and 4). The locations of nDEP and pDEP electrostatic traps on the JP surface were obtained from three-dimensional (3D) numerical simulations (Fig. S3). Notably, pDEP trapping of capsules at the dielectric hemisphere's equator was observed at solution concentrations <1 mM NaCl, while nDEP trapping at their metallic side was observed at solution concentrations >1 mM NaCl (Fig. 2c). At solution concentrations below 1 mM NaCl, the average number of pDEP-trapped capsules per JP was one or two when combining an electric field (at all frequencies) with magnetic rolling, as the capsules were sheared off by magnetic rolling, preventing them from either trapping or remaining on the surface of JP (Fig. S4a and b).

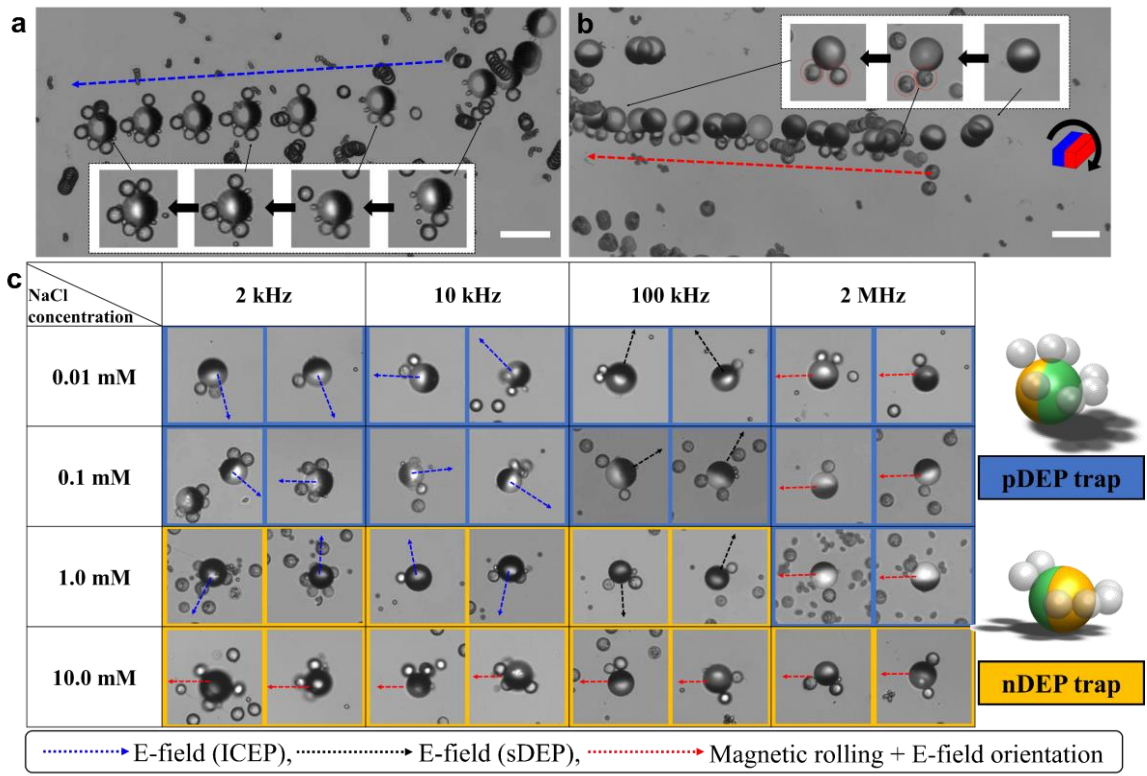

**Figure 2. Electrical trapping of jellyfish capsules on Janus particles and magnetic field-based transportation of the assembled biohybrid microbots.** (a) Representative transport of a biohybrid microbot (MR) carrying jellyfish capsules within 0.1 mM NaCl, subjected to a unified external AC electric field (2 kHz, 15 $V_{pp}$), without magnetic field, to drive self-propulsion and capsule trapping. (b) The MR was subjected to an external rotating magnetic field (100 rpm) along with an AC electric field (2 MHz, 15$V_{pp}$) for self-propulsion and capsule trapping within 10 mM NaCl, respectively. (c) Gallery of the assembled MRs with their transport directions at varying conductivities of NaCl and applied electric field frequencies. Two 3D schematic description of the possible trapping locations of the capsules onto the MRs under positive dielectrophoresis (pDEP) (navy-blue color filled rectangles) and negative dielectrophoresis (nDEP) (yellow-orange color filled rectangles) trapping conditions. Blue, black and red dashed arrows indicate the transport direction driven by electric fields, for induced-charge electro-phoresis (ICEP) (dielectric hemisphere forward), self-dielectrophoresis (sDEP) (metallic-coated hemisphere forward) and magnetic rolling with electric field orientation, respectively.

In high-conductivity solutions (10 mM NaCl) or at frequencies exceeding 1 MHz, the JP is exclusively propelled through magnetic rolling, as the electric field ceases to contribute to propulsion. Instead, the electric field plays a role in orientating the JP, to improve the stability of its rotation along the axis perpendicular to its metallo-dielectric interface, and to facilitate the DEP trapping of the jellyfish capsules on the JPs surface (see Fig. 2b, Fig S1 and movie S1).Comparing to lower concentration below 1mM NaCl, more capsules were trapped onto the JP surface (Fig.S4c), due to the decreased electric double layer (EDL) resulting in electrostatic absorption. Following trapping, the capsules seemed to remain adsorbed to the JP surface in the absence of AC electric fields. It was also observed that JPs tend to form clusters due to both magnetic and electrostatic interactions, enhancing their capacity to trap capsules (Fig. S4c). The optimal cluster size appears to be three to six JPs, as these clusters demonstrate high efficiency in transport (fast velocity) and capturing capsules, with a maximum loading capacity of approximately 10 capsules. However, clusters with more than 10 JPs encounter transport difficulties due to their large volume, and the number of captured capsules decreases because the surface area for trapping becomes blocked. High-conductivity solutions of up to 10mM NaCl were tested, yet the range should be extended to close to physiological solution conditions such as 100 mM NaCl (data not shown), cell growth medium, 10 mM PBS, seen in fig.1b). These solutions exhibit consistent DEP trapping and propulsion behavior similar to that observed with 10 mM NaCl and are suitable for biosamples, e.g. cancer spheroids, as they maintain the osmolarity balance[40].

**Transportation and activation of biohybrid MRs**

The transportation and activation of the assembled biohybrid MR was illustrated using a simplified microfluidic setup adapted from our previous work[33] (see Fig. S5a). A mixture of 27 μm-sized JPs and jellyfish capsules was introduced into a microchamber filled with physiological buffer (σ ~15 mS/cm), through one of the drilled inlet holes. They were then assembled and transported to the target region located ~7 mm away from the inlet, using a combination of rotating magnetic and electric fields. Subsequently, 1 % w/v subtilisin was introduced through the other inlet hole to trigger rapid tubule ejection and delivery of the preloaded molecular contents of the capsules (Fig. 3).

Fig. 3a depicts various biohybrid MRs, including single, dual and triple JP configurations, each loaded with a different number of intact capsules within 10mM NaCl (see movie S3). An AC electric field with frequencies of 100 kHz and 2 MHz was then applied to the first two and the last MR, respectively. During magnetic rotation, the capsules assembled on the JPs exhibited minimal detachment from the surface. The presence of assembled MRs even before applying the electric field as well as the attachment of the capsules onto the entire surface of JPs, including onto the equator of the metallic-coated hemisphere, as demonstrated by nDEP trapping, provide additional evidence of non-specific adsorption of capsules to the JPs. This phenomenon leads to less detachment of capsules by magnetic rotation-induced hydrodynamic shearing. At the activation stage, the diffused enzymes triggered the inside-out eversion of tubules within most of the capsules in the biohybrid MR. This process led to rapid elongation of the tubules, reaching an average length of 298 ± 20 μm. While the experimental setup and camera speed were not sufficient for monitoring the speed of tubule ejection, previous studies indicated velocities between 10 mm s$^{-1}$ and 10 m s$^{-1}$[17,41]. The release mechanism depends on the level of enzyme exposure to the capsules. On average, it took approximately 5 minutes to activate over 70% of the capsules compared to the total counted before activation in an open chamber. However, in the microfluidic chamber (Fig. S6, case A and B), there was an additional delay primarily due to the time required for the enzyme to diffuse and reach the specific locations (1 mm distance from the left-inlet hole) where the biohybrid MRs had been transported (Fig. S6). Due to the uncontrolled stochastic direction of tubule ejection, a large number of MRs carrying many jellyfish capsules is suggested to be an efficient solution for effective drug delivery. Ejection of toluidine blue and acridine orange molecules, preloaded within the capsules was observed (Figs. 3b and c, respectively). Movies S4 to S5 and Fig. S5b show the entire MR transportation process and activation. Most molecules were ejected through the tubule tip rather than along the tubule, and the capsules were empty after ejection. All molecules, except for those adsorbed, were ejected from the capsules once the tubules were fully elongated following activation, which occurred after approximately 11 minutes of enzyme exposure (Fig S6c). Interestingly, some capsules were disconnected from the JP surface following the discharge event, likely due to the rebound-like forces associated with the discharge. These results demonstrate the ability to control the transport of assembled biohybrid MRs to specific locations and to activate rapid tubule elongation of ~300 μm and subsequent ejection of the capsules' molecular contents. This innovative approach holds promise for delivering drugs deep into organs and tissues.

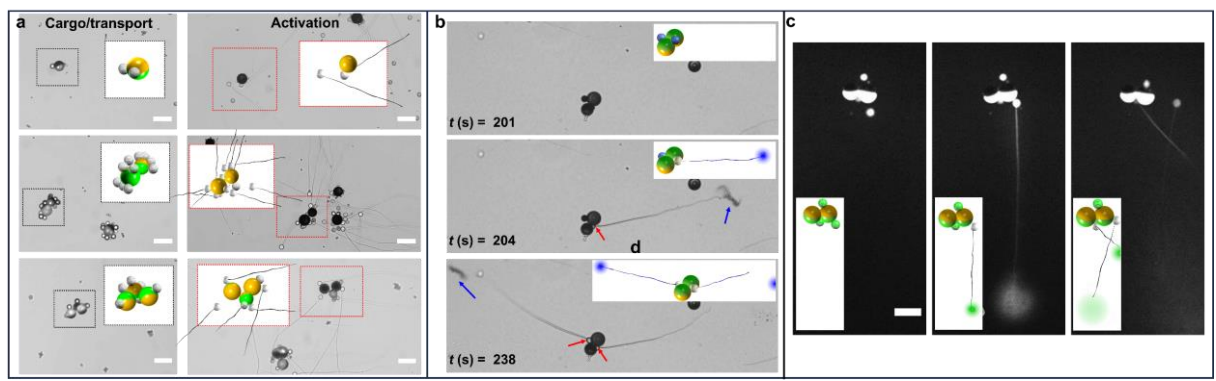

**Figure 3. Enzyme-triggered tubule elongation of transported jellyfish capsules results in ejection of their preloaded molecular content.** (a) Different compositions of Janus particles (JPs), comprising single, dual, and triple JPs, are transported and exposed to an enzyme-rich environment, which leads to the activation of the intact capsules trapped onto the microbots. Insets illustrate schematics of the corresponding microrobot states. (b) Time-lapse microscopic images showing the ejection process of preloaded toluidine blue O dye from biohybrid microbots. Blue arrows indicate the release of molecules from the capsule bodies through tubules. Empty capsule bodies are highlighted by red arrows. (c) Fluorescence microscopy images showing ejection of fluorescently-tagged acridine orange from the tip of the tubule (scale bar: 50 μm).

**In vitro deep penetration of biohybrid microbots tubules into cancer spheroids**

To assess the potential used of biohybrid MRs for deep tissue penetration, in-vitro experiments using cancer spheroids derived from the 293T cancer cell line, were conducted within a microfluidic chamber containing cell culture medium (σ ~18 mS/cm) (Fig. 4). A MR swarm was transported from the right chamber holes (Fig. 4b), over a distance of ~8 mm, to 1 mm-diameter spheroids placed near the left chamber hole (movie S6). Upon enzymatic activation of the capsules, the ejected tubules deeply penetrated and injected their pre-loaded molecular content into the targeted spheroids.

Time-lapse images depicting the ejection of preloaded molecules, including toluidine blue and acridine orange, into the spheroids are shown in Figs. 4c and d and are complemented by movies S6 and S7, respectively. The initial time (t = 0 s) was arbitrarily set one frame before tubule activation. In the case of toluidine blue, the injection of the molecules into the cancer spheroid is seen by the region with lower image brightness associated with the ejected molecules, illustrating the successful injection of molecules through the tip of the ejected tubule to a depth of ~300 μm from the edge of the spheroid. In the case of acridine orange, a clear fluorescent trace was observable along the ejected tubule and within the spheroid (Fig. 4d), likely due to its perforated structure[17,42]. To minimize enzyme toxicity[43], its concentration was maintained at a maximum of 0.2% w/v. Upon activation, the tubules fully extend, completely unfolding from within the capsules. Nonetheless, the cells on the outer surface of the spheroid displayed significant swelling and morphological changes upon diffusion of the natural jellyfish toxin (e.g., polyclonal antitoxin $A_2$ [44]) from nearby activated capsules that did not penetrate the spheroid due to the stochastic direction of tubule ejection. Overall, the percentage of tubules that penetrated the spheroid, relative to the total number ejected, was approximately 23% based on multiple trials. To increase the chances of effective spheroid penetration, the number of biohybrid MRs transported to the targeted spheroid was increased.

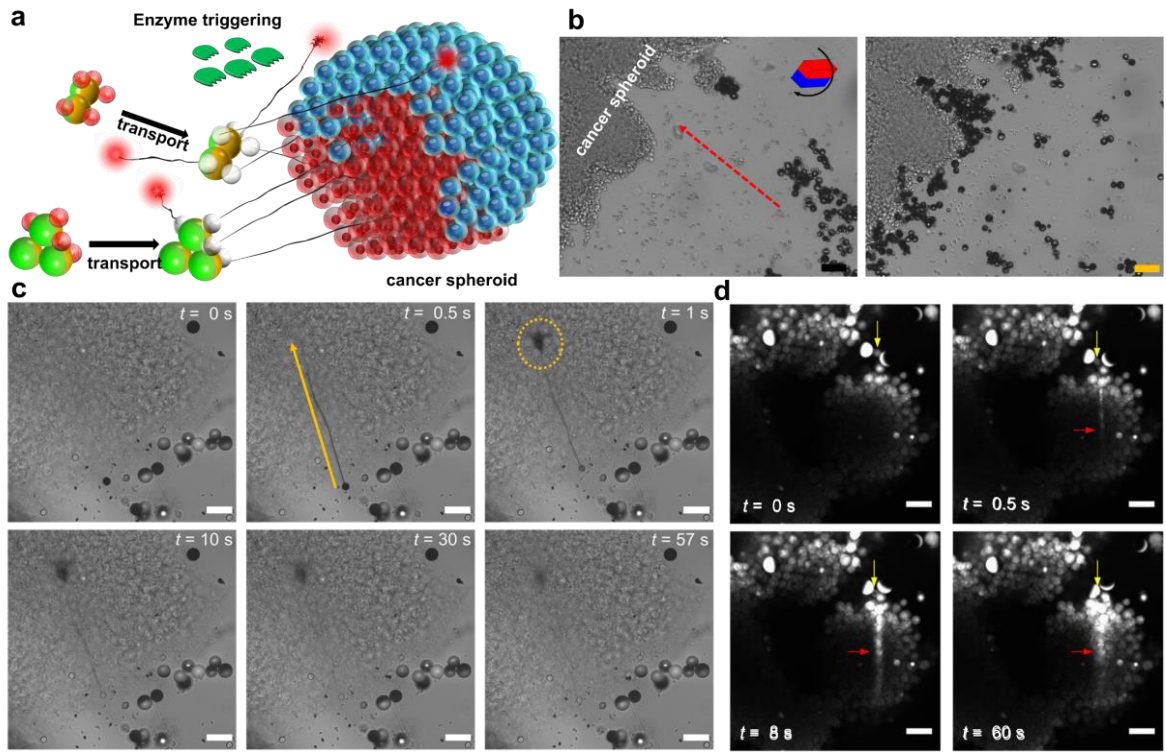

**Figure 4. Deep penetration of biohybrid microbots tubules into a cancer spheroid.** a) A schematic of the deep penetration of preloaded molecular content from the capsules of a biohybrid microbot (MR) into cancer cell spheroids, achieved through the transport of the biohybrid microbots followed by activation of the loaded capsules by a specific enzyme. (b) Microscopy images depicting transport of a swarm of MRs toward a target spheroid (direction is indicated by a red arrow) using magnetic rolling with an electric field (15 $V_{pp}$, 2 MHz) (scale bar:

100 μm.) (c) Time-lapse microscopy images capturing penetration of the ejected tubule followed by injection of the preloaded content (toluidine blue O) deep into the spheroid. The orange arrow indicates tubule elongation from the jellyfish's capsule body and the dashed circle marks the ejected dye. (d) Time-lapse fluorescence microscopy images showing tubule penetration into the spheroid (indicated by a yellow arrow) and ejection of fluorescently-tagged molecules along the tubule (indicated by a red arrow) (scale bar: 50 μm).

**Penetration of biohybrid microbot tubules into living *Caenorhabditis elegans***

*Caenorhabditis elegans* (*C. elegans*) was employed as a target animal model to investigate the penetration capabilities of the biohybrid MRs (Fig. 5). It should be noted that *C. elegans* serves as an excellent animal model for studying the effects of neural toxins due to its simple and well-characterized nervous system[45]. Unlike the conventional microinjection method[46,47] involving glass capillaries and halocarbon oil-mounted worms on an injection pad, the present setup utilized freely swimming, age-synchronized *C. elegans* (~250 μm in length and ~15 μm in diameter). The worms were introduced into the left side of a microfluidic chamber filled with worm growth medium. The biohybrid MRs, injected through the right hole, propelled toward the swimming *C. elegans* by magnetic rolling under an electric field (15Vpp, 2 MHz). Once they reached the target, an enzyme (0.3% w/v subtilisin) was introduced for capsule activation. Of note, mixtures of different types of capsules with different dimensions (i.e., small and large capsules [48]) were assembled together onto the JPs. The assembled MRs, composed of JP clusters with several (~10) loaded capsules containing blue dyes, were transported far from the inlets, chasing healthy C. elegans and attempting to penetrate the swimming worms. However, they neither penetrated the worms nor elicited a significant reaction (Movie S8). To improve penetration, the mobility of the worms was decreased through paralysis, by addition of 20 μM levamisole.

Figures 5b and c, complemented by Supplementary Movie S9, illustrate the sequential events focusing on capsule shooting at the worms and their subsequent motion response. It is noted that the C. elegans were located near the right chamber hole, where untrapped jellyfish capsules were also randomly dispersed. For instance, for worm "1," marked with a yellow dot circle in Fig. 5b and inset in Fig. 5b, initial motion nearly ceased before capsule activation. Following activation and penetration of the small capsule tubule, the worm transitioned from a U-shape to a snake-coiled configuration within 2 minutes. This behavior resembles an escape response[49,50], potentially serving to protect its bodies from potentially threatening stimuli, such as toxins, and thermal stimuli. Interestingly, upon tubule penetration from large capsules, the paralyzed worm exhibited a sudden vigorous motion that lasted for ~ 4.5 minutes (Fig. 5c and Supplementary Movie S9). In another case of a non-paralyzed worm, cessation of movement was observed upon penetration of the tubule of a small capsule (Movie S10d). It was noted that the abrupt reactions of *C. elegans* were not triggered by other potential stimuli, such as the cytotoxicity of the introduced enzyme, the applied external electric field, or magnetic rolling, as evidenced by the control tests (movie S10a to c). It is hypothesized that a muscle twitching effect was responsible for the sharp movements, also known as omega turns[50,51], upon tubule penetration, likely triggered by exposure to the natural jellyfish toxins[52] and not necessarily due to the mechanical penetration itself of the thin tubule (~1 μm in diameter[24]). Further investigation is needed to gain a deeper understanding of post-penetration responses, including potential neurotoxic effects[52].

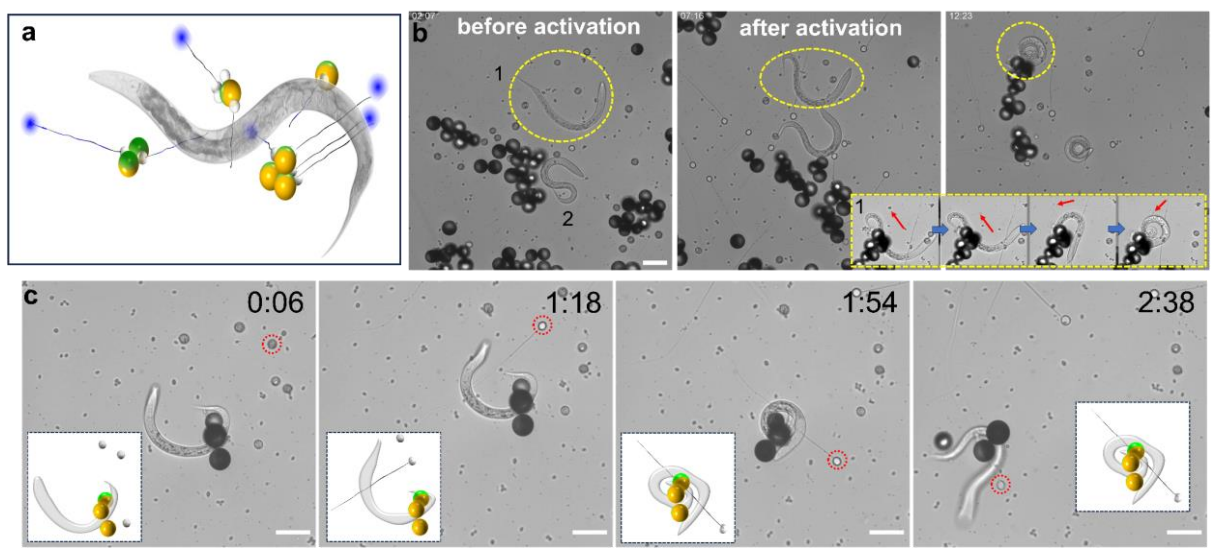

**Figure 5. Biohybrid microbots penetrating living *Caenorhabditis elegans*** (a) A schematic of the penetration of tubules of jellyfish capsules in biohybrid microbots into live *C. elegans*. (b) Time-lapse microscopy images capturing the moments before and after capsule activation and the corresponding worm reaction. The yellow-dotted inset provides a close-up view of the body bending of worm "1" (yellow circle), induced by penetration of small capsules (red arrows). (c) Time-lapse microscopy images capturing the penetration of tubules from a large capsule into paralyzed *C. elegans*, resulting in abrupt motion and noticeable body bending. Scale bar: 50 μm.

## Conclusions

This study introduced a novel nematocyst-based biohybrid MR system that enabled drug delivery deep into targeted tissues in in-vitro assays. The MRs were assembled through DEP-assisted trapping, and transported via magnetic rolling-based propulsion, assisted with electric field-based orientation so as to navigate them to a targeted region of interest. After reaching the target destination, capsule activation was triggered with a specific enzyme. Deep penetration of the ejected tubule into cancer spheroids and living organisms, such as *C. elegans*, was demonstrated, and highlighted the potential of this technology for targeted drug delivery applications. In the case of *C. elegans*, interesting post-injection responses, including possible neurotoxic effects on *C. elegans* muscles, were observed and warrant further investigation.

Future developments of this unique system will include replacement of the toxic contents of the capsules with various drugs of therapeutic value, as well as use of different types of capsules to enhance versatility and to improve biocompatibility[53]. Future developments may also include realization of bioinspired engineered synthetic MRs preloaded with drugs and designed with various triggering and ejection mechanisms (e.g., osmotic pressure[19]) for their release. The controlled and precise navigation of the biohybrid MRs demonstrated here within an in-vitro microfluidic platform, serves as a foundation for their further exploration in more complex biological environments, as well in in-vivo applications. Although the reduced concentration of subtilisin protease (0.2%) in in-vitro tests resulted in minimal cytotoxicity over a few hours[54], future research could explore encapsulating enzymes with effective in vivo transport and release strategies[55,56] using various synthetic and biopolymers - such as protein capsids, hydrogels and lipid-based particles[57] - outside the stinging capsule or developing alternative external triggering mechanisms (e.g. acoustic) to induce capsule discharge that are safely for potential in-vivo applications, eliminating the need for enzymes. Such precise MRs that move within complex biological environments and enable controlled delivery of drugs deep into targeted tissues, hold great potential in advancing precise drug delivery for various biomedical applications. While our approach is still in its early stages and not yet suitable for in vivo applications, it shows promise for assessing deep tissue drug delivery effects in in vitro models, such as organ-on-a-chip systems that incorporate organoids, organelles, or dissected tissues. This research provides insights into drug penetration beyond the tissue surface [58, 59] and may inspire future studies focused on in vivo deep tissue drug delivery applications.

## Materials and Methods

**Preparation of Janus particles and experimental setup**

Magneto-metallo-dielectric JPs (27 μm in diameter) were fabricated by coating fluorescent polystyrene particles (Sigma Aldrich) with 15nm Cr, followed by deposition of 50 nm Ni and 15 nm Au layers using an electron-beam evaporator, as previously described[33]. Before releasing the JPs from the glass slides, they were magnetized by placing the slide in between two neodymium magnetic blocks, with opposite dipoles aligned parallel to the metallo-dielectric interface of the JP[37]. The JPs were released by sonication in deionized (DI) water, and then rinsed three times with DI water before their introduction into the final solutions.

The single microchamber device comprised a circular microfluidic chamber, with a spacer positioned between two indium tin oxide (ITO)-coated glass slides (Delta Technologies), as previously described[33]. The bottom ITO-coated glass slide underwent an additional coating with 20 nm-thick silicon dioxide using a sputter (AJA international Inc., ATC 2200), to minimize particle adsorption onto the substrate. The thin spacer, which formed a 9 mm-diameter microchamber, was made from 100 μm-thick double-sided tape (3M), cut with an electronic cutting machine (Silhouette Cameo 3, Silhouette America Inc.).

Various alternating current (AC) frequencies with a sinusoidal wave form were applied using a function generator (33250A, Agilent) for the dielectrophoretic attractions-based assembly of MRs and jellyfish capsules and to promote the magnetic rolling of the assembled MRs. To facilitate transportation of assembled biohybrid MRs, an external rotating magnetic field was applied via a homebuilt system consisting of a fixed neodymium magnet (grade: N35) mounted on a motor for its controlled rotation with a corresponding control unit. Transport and activation of biohybrid microrobots were captured using an Andor Neo sCMOS camera mounted on an inverted epi-fluorescence microscope (Eclipse Ti-U, Nikon) or a spinning disc confocal microscopy system (Yokogawa CSU-X1) connected to an inverted microscope (Eclipse Ti-U, Nikon) and a camera (Andor iXon3) equipped with a 10×/20× objective lens for obtaining bright-field and fluorescence microscopy images, respectively. Motion analysis of biohybrid microrobots and quantification of fluorescent cargo intensities after binding events, were conducted using ImageJ software.

**Jellyfish capsule preparation and staining**

Jellyfish (isorhiza nematocysts) capsules were isolated from the tentacles of *Rhopilema nomadica* jellyfish[44] collected in Haifa Bay, Israel, as described by Rachamim et al[14]. The tentacles were briefly homogenized and separated in a 50% Percoll (GEHealthcare) gradient before being washed with distilled deionized water (DDW). Of note, jellyfish capsules of heterogeneous morphologies (i.e., different sizes and shapes) were isolated[48]. Capsules were re-suspended in 10 mM NaCl and stored at 4 °C for one day before the experiments. For molecule ejection experiments, 0.1% toluidine Blue O and 0.1% acridine orange hemi(zinc chloride) salt were loaded into the capsule after mixing capsules for 5 min. The stained capsules were then centrifuged at 4 °C, 800 RCF, for 4 min, resuspended in PBS, cell culture medium, or worm growth medium, and then introduced together with JPs into one of the chamber's inlets.

**Cancer spheroid preparation**

The 293T cancer cell line was cultured at 37 °C, 5% $CO_2$, in Dulbecco's modified Eagle medium (DMEM, Biological Industries), supplemented with 10% v/v heat-inactivated fetal bovine serum and 1% v/v penicillin-streptomycin (Biological Industries). Cells were passaged when reaching 90% confluency, typically every 3 days. The conventional hanging drop method was used to generate 3D cancer cell spheroids[60] one day before introducing the cells into the microchamber device. The resulting cell spheroids were gently introduced into the left side of the chamber, which was filled with cell culture medium.

*C. elegans* **preparation and paralysis**

The N2 Bristol strain of *Caenorhabditis elegans* (*C. elegans*) was provided by Prof. Gili Bisker, Tel-Aviv University, Israel. The worms were cultured on standard nematode growth medium (NGM) plates, which were supplemented with *Escherichia coli* strain OP50 as a food source, and maintained at a temperature of 20 ± 1 °C. To ensure the vitality of the stock culture, the worms were transferred to fresh NGM plates every 3-4 days, following a protocol previously described[45,61].

# Acknowledgements

G.Y. acknowledges support from the Israel Science Foundation (ISF) (1934/20). We would like to thank Dr. Adi Hendler and Prof. Gili Bisker from Tel-Aviv University for their help in the *C. elgans* study and Prof. Limor Broday from Tel-Aviv University for her advice with the *C. elgans* experiments.

# Supplementary information

# Biohybrid Microrobots Based on Jellyfish Stinging Capsules and Janus Particles for In Vitro Deep-Tissue Drug Penetration

Sinwook Park[1,3], Noga Barak[2], Tamar Lotan[2], and Gilad Yossifon[1,3*]

[1]School of Mechanical Engineering, Tel-Aviv University, Tel Aviv, 6997801, Israel

[2]Marine Biology Department, The Leon H. Charney School of Marine Sciences, University of Haifa, Haifa, 310330, Israel

[3]Department of Biomedical Engineering, Tel-Aviv University, Tel Aviv, 6997801, Israel

* Corresponding author: gyossifon@tauex.tau.ac.il

**Movie S1:** Two movies show the transport of a biohybrid microbot carrying jellyfish capsules: one driven by a unified external AC electric field (2 kHz, 15$V_{pp}$) within 0.1 mM NaCl without a magnetic field, and the other by a combined external rotating magnetic field (rotation speed with 100 rpm) and an AC electric field (2 MHz, 15$V_{pp}$) within 10 mM NaCl, corresponding to Figs. 2a and 2b, respectively.**Movie S2:** Self-propulsion behavior of JPs in different concentrations of NaCl solution under the application of an electric field with varying frequencies (2 kHz, 10 kHz, 50 kHz, and 2 MHz) and magnetic rolling.**Movie S3:** Combined two movies showing the transport and activation of the loaded jellyfish capsules by enzyme triggering, focused on biohybrid microrobots with various configuration of JPs corresponding to Fig. 3a.

**Movie S4:** Transport and molecule ejection of Toluidine Blue O, (TBO) in 10mM NaCl solution corresponding to Fig. 3b.

**Movie S5:** Transport and molecule ejection of Acridine Orange in 10mM NaCl solution corresponding to Fig. 3c.

**Movie S6:** Transport of the microbots towards the target spheroid using magnetic rolling assisted with electric field corresponding Fig. 4b and representative event of molecule injection of Toluidine Blue O, TBO deep into the spheroid corresponding to Fig. 4c.

**Movie S7:** Molecule injection of Acridine Orange hemi(zinc chloride) salt deep into the spheroid from the loaded jellyfish capsules of the biohybrid microrobot by enzyme triggering corresponding to Fig. 4d.

**Movie S8:** Transport of biohybrid microrobots toward swimming healthy C. elegans and their unsuccessful attempts to penetrate capsule tubules into the target worms due to the latter's fast mobility.**Movie S9:** The penetration of the biohybrid microrobots' tubules from the small and large capsule into paralyzed *C. elegans* by enzyme triggering corresponding to Figs. 5b and c, respectively.

**Movie S10:** Combined movies of control tests to observe the reactions of C. elegans under various potential stimuli: a) Control test with only enzyme introduction without an external electric field. b) Control test introducing both the enzyme and applying an external electric field with magnetic rolling,

using paralyzed worms. c) Control test without the penetration of tubules under conditions of electric/magnetic fields and enzyme introduction. d) Test showing the penetration of tubules from the small capsule into non-paralyzed C. elegans by enzyme triggering.

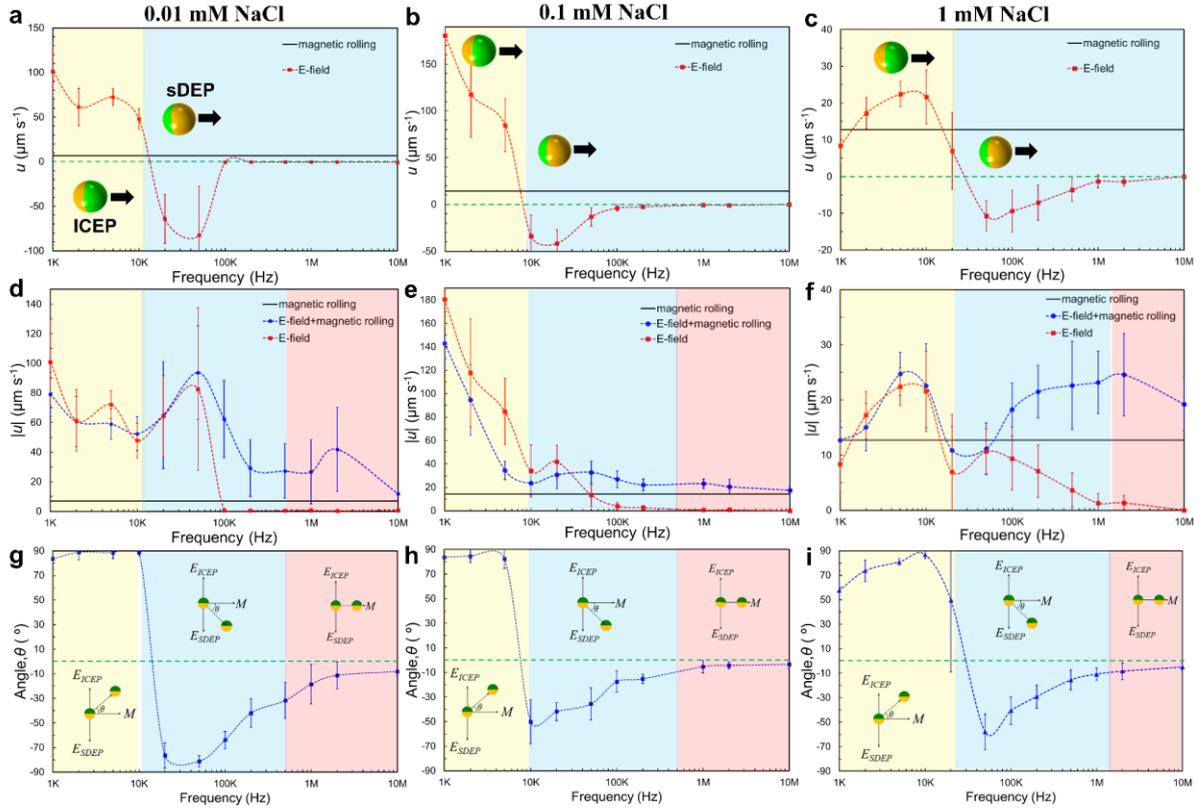

**Figure S1. Propulsion behavior of Janus particles in different concentrations of NaCl solution under the application of an electric field and magnetic rolling.** (a, b, c) Velocity profiles of Janus particle self-propulsion under decoupled magnetic rolling and electric field application within NaCl concentrations of 0.01, 0.1, and 1 mM, respectively. (d, e, f) Absolute velocity magnitudes under three conditions: only magnetic rolling, coupled magnetic rolling and electric field, and only electric field, within NaCl concentrations of 0.01, 0.1, and 1 mM, respectively. (Refer to representative movie S2 as an example of Janus particle movement.) (g, h, i) Changes in the angle (θ) of the direction of motion relative to the direction of only magnetic rolling (set as the horizontal axis) when both the electric field at various frequencies and magnetic rolling are coupled. Yellow, green, and red rectangles indicate dominance of ICEP, sDEP, and magnetic rolling transport of Janus particles, respectively, upon the applied frequency of the electric field.

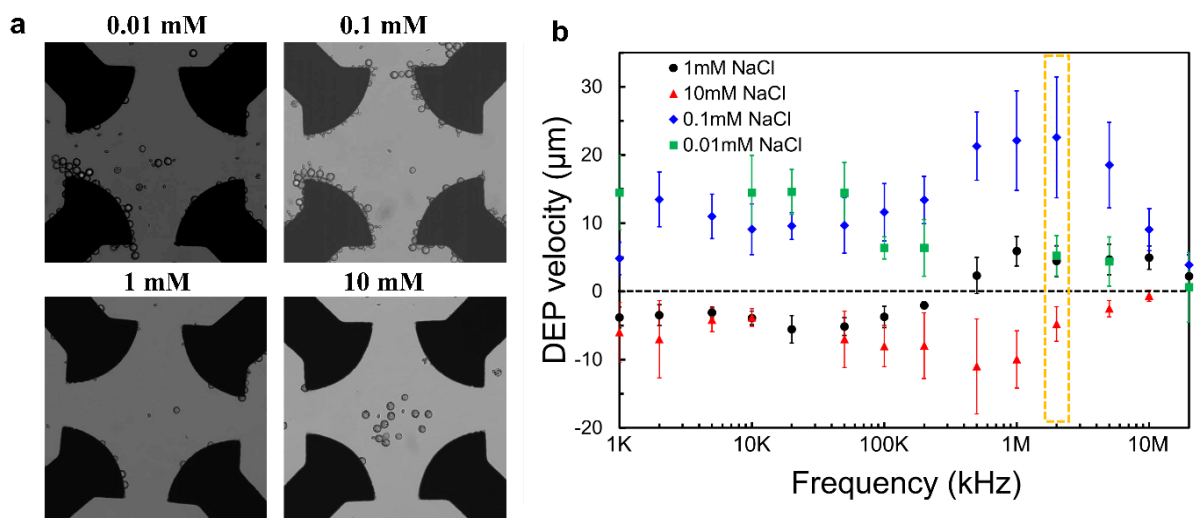

**Figure S2. DEP characterization of the jellyfish capsules under different concentrations of NaCl solution.** (a) Microscopic images illustrating pDEP and nDEP behaviors of the capsules at NaCl concentrations of 0.01, 0.1, and 1 mM, and 10 mM NaCl, under an applied electric field (2 MHz, 10$V_{pp}$). (b) Experimental DEP spectra of jellyfish capsules in various NaCl concentrations. The orange dashed rectangle highlights distinct DEP behaviors of jellyfish capsules at 2 MHz corresponding to part (a).

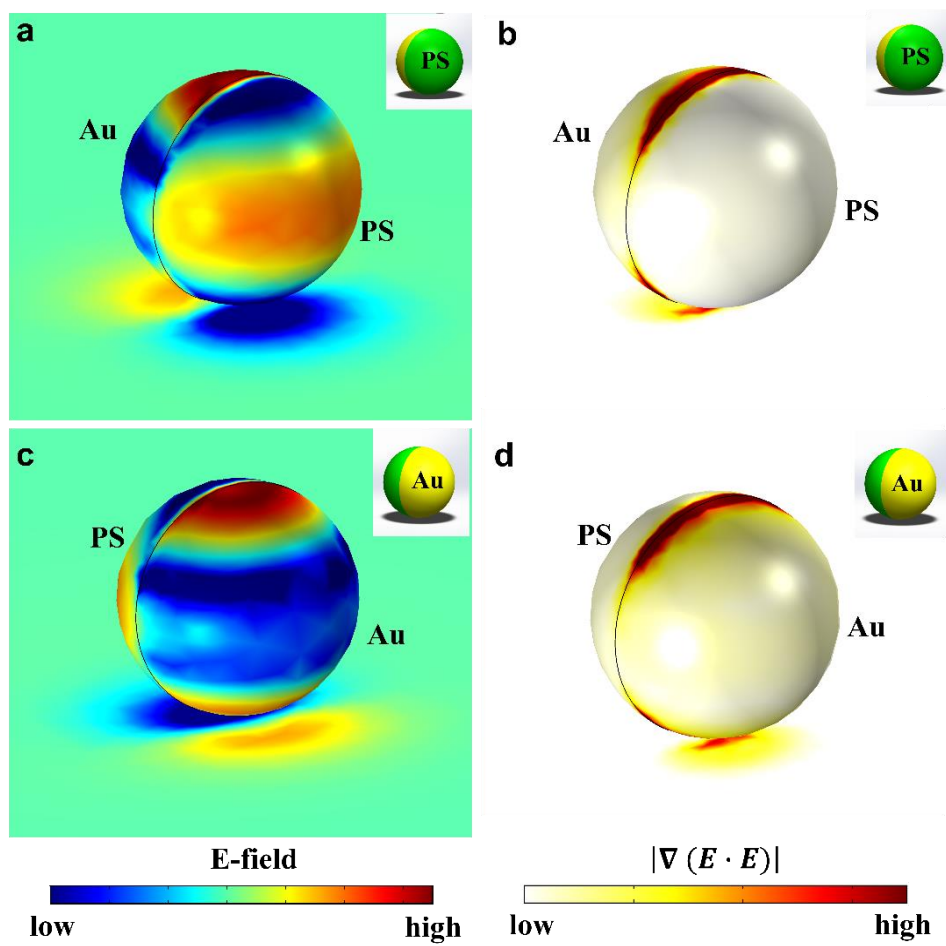

**Figure S3. 3D simulation of the electric field norm and its gradient onto the 27μm metallo-dielectric Janus particles.** (a, c) electric field norm (b, d) its gradient on the dielectric and metallic surfaces, respectively.

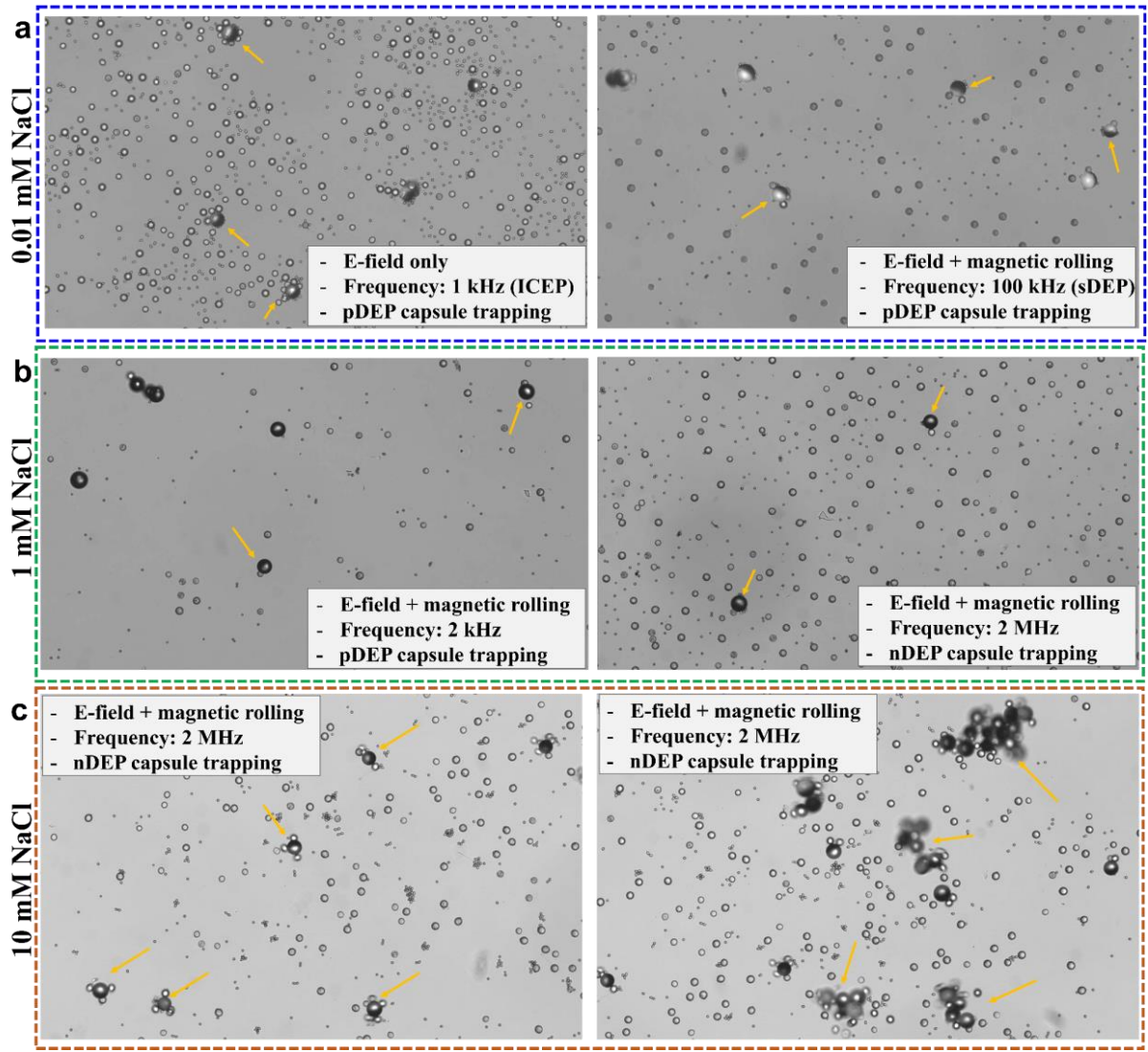

**Figure S4. Transport behavior of assembled biohybrid microrobots under varying frequencies of the applied external electric field and different NaCl concentrations using 27μm metallo-dielectric JPs.** (a) Comparison of the propulsion and cargo loading capacity using pDEP trapping of capsules between only electric field application with ICEP propulsion mode and a combination of magnetic rolling and electric field application with sDEP propulsion mode wtihin 0.1 mM NaCl. (b) Comparison of pDEP and nDEP capsule trapping under the applied electric field frequencies of 2kHz and 2 MHz, respectively, within 1 mM NaCl. (c) Comparison of individual JPs and clusters of JPs based assembly with capsules within 10mM NaCl.

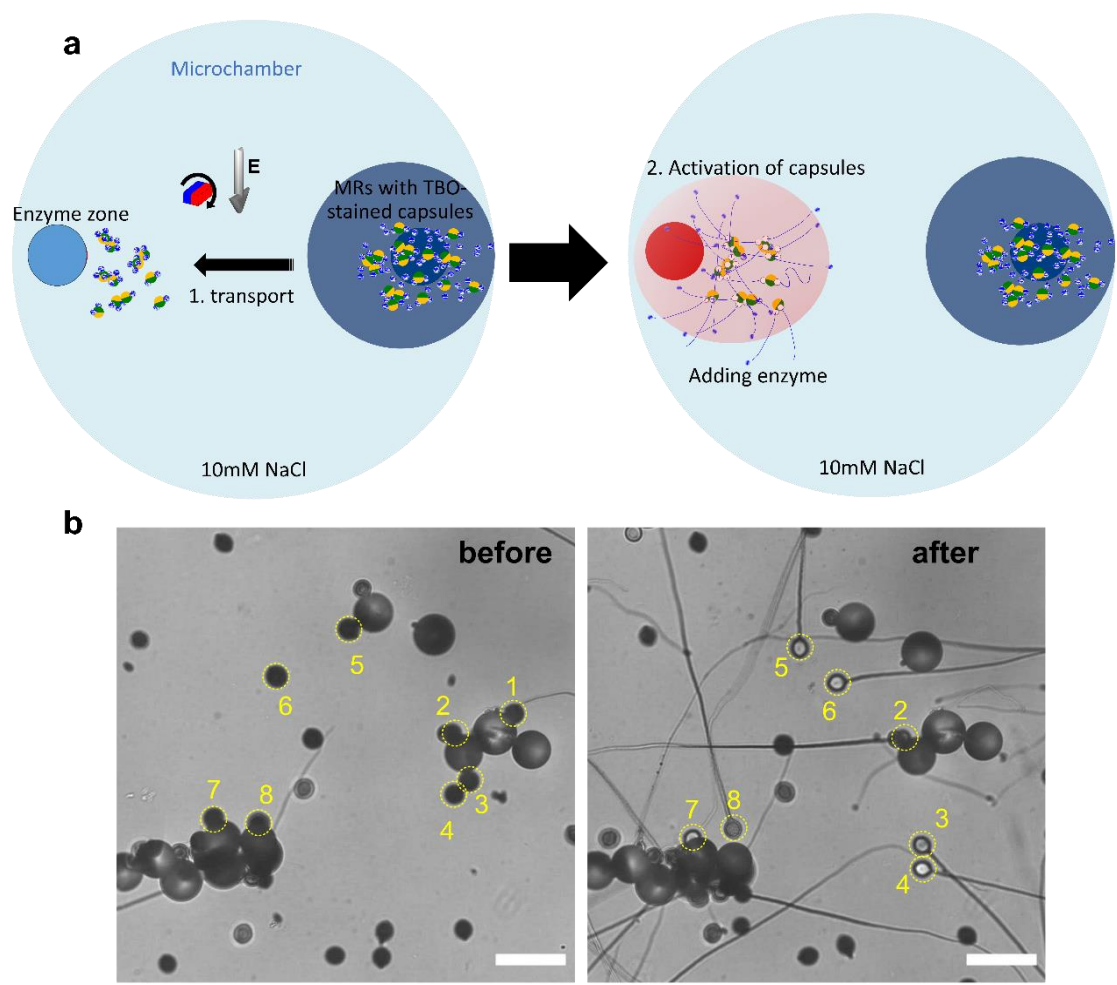

**Figure S5. Molecule ejections from biohybrid microrobots by enzyme triggering of the jellyfish capsules.** (a) Schematics of the molecule ejection procedure, illustrating (i) the transportation of assembled biohybrid microbots to the left side of the microchamber, and (ii) activation of the loaded capsules by introducing enzyme. (b) Microscopic images depicting biohybrid microbots before and after activation. Yellow dot circles with numbers indicate the marked jellyfish capsules before and after activation, highlighting the release of preloaded Toluidine Blue O molecules from the capsule through their released tubules.

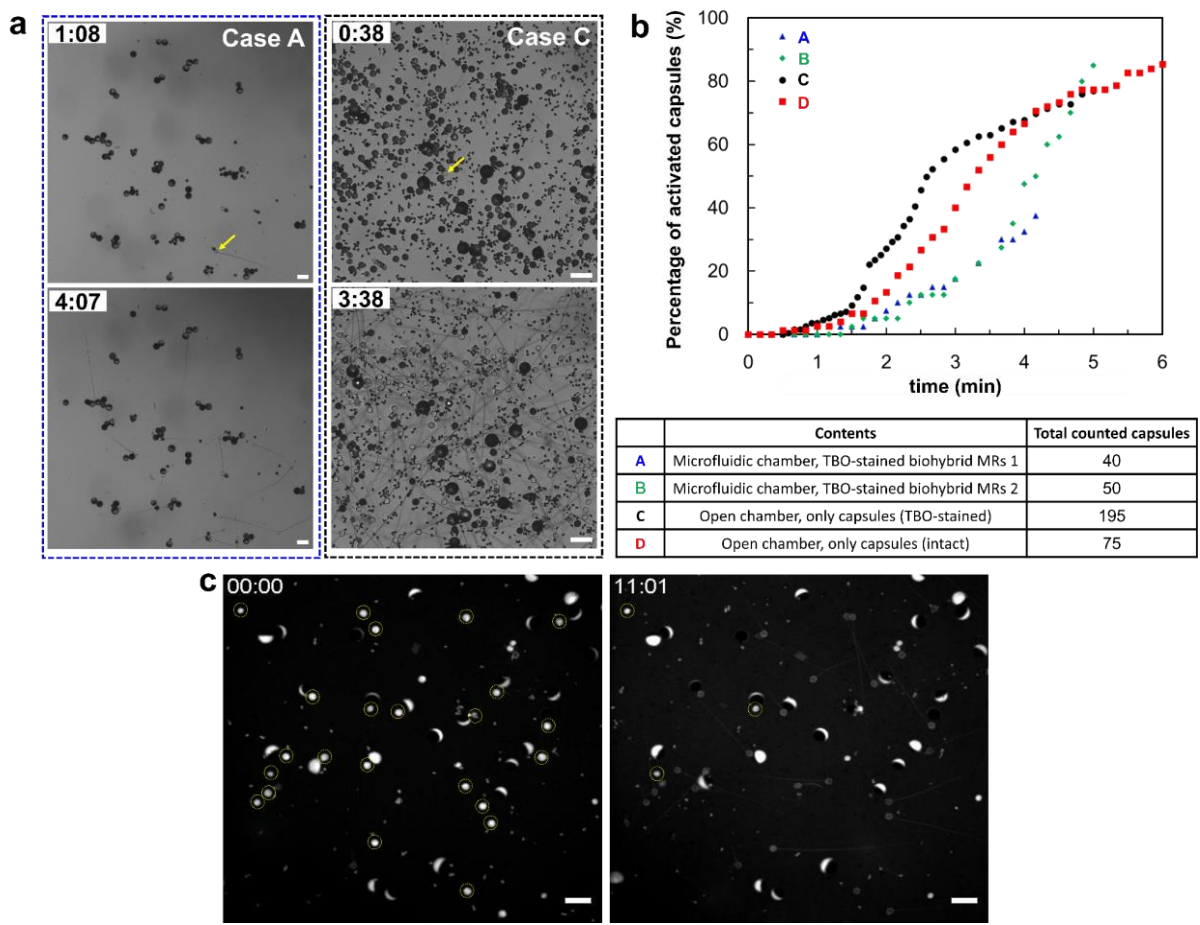

**Figure S6.** Time evolution of molecule-preloaded capsule activation and ejection. (a) Representative time-lapse microscopic images showing the first tubule activation (t = 68 s and 38 s) after enzyme introduction (t = 0 s) and 3 minutes after first activation (t = 247 s and 218 s) for biohybrid MRs with toluidine blue O (TBO)-preloaded capsules (A, blue dashed line) and TBO-preloaded capsules alone (C, black dashed line), respectively. (b) Corresponding time-evolution graph illustrating the percentage of activated capsules with ejected tubules relative to the total capsules counted before activation across four conditions: A, B – transport of TBO-preloaded biohybrid MRs within a microfluidic chamber (~1 mm from the left inlet); C – TBO-preloaded capsules alone in an open chamber; D – intact capsules (without encapsulated molecules) in an open chamber. (c) Representative time-lapse microscopic images showing capsule activation from biohybrid MRs preloaded with the fluorescent molecule acridine orange hemi(zinc chloride) salt after sufficient enzyme exposure (11 minutes) within a microfluidic chamber.

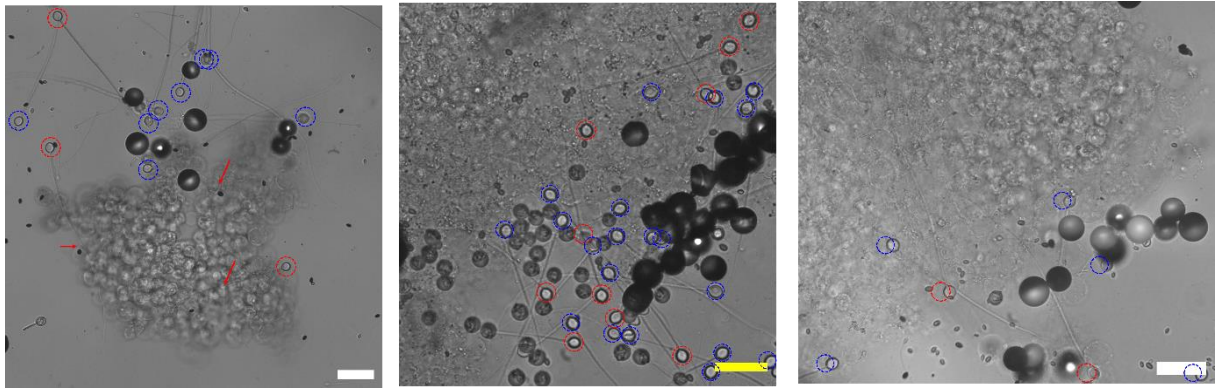

**Figure S7**. Percentage of tubule penetration into a cancer spheroid by activated capsules. Red dashed circles indicate successful penetration, while blue dashed circles highlight unsuccessful attempts. The spheroid volumes are 0.006, 0.007, and 0.01 mm³ (left to right).